\documentclass[conference]{IEEEtran}
\usepackage{array}
\usepackage{cite}
\usepackage{amsmath,amssymb,amsfonts}
\usepackage{algorithmic}
\usepackage{graphicx}
\usepackage{booktabs}
\usepackage{ragged2e}
\usepackage[caption=false,font=normalsize,labelfont=sf,textfont=sf]{subfig}
\usepackage{textcomp}
\usepackage{threeparttable}
\usepackage{multirow}
\usepackage{stfloats}
\usepackage{url}
\usepackage{textcomp}
\usepackage{xcolor}
\def\BibTeX{{\rm B\kern-.05em{\sc i\kern-.025em b}\kern-.08em
    T\kern-.1667em\lower.7ex\hbox{E}\kern-.125emX}}
    
\begin{document}
\onecolumn
\begingroup
\fontsize{20pt}{20pt}\selectfont
IEEE Copyright Notice\\
\endgroup
\\
© 2025 IEEE. Personal use of this material is permitted. Permission from IEEE must be obtained for all other uses, in any current or future media, including reprinting/republishing this material for advertising or promotional purposes, creating new collective works, for resale or redistribution to servers or lists, or reuse of any copyrighted component of this work in other works.
\twocolumn

\title{Practical Equivalence Testing and Its Application in Synthetic Pre-Crash Scenario Validation
\thanks{This work was supported by the FFI program sponsored by Vinnova, the Swedish Energy Agency, and the Swedish Traffic Authority as part of the project Improved quantitative driver behavior models and safety assessment methods for ADAS and AD (QUADRIS: nr. 2020-05156).
The work was also partially sponsored by the V4SAFETY project, funded by the European Commission under grant number 101075068.}
}

\author{\IEEEauthorblockN{Jian Wu}
\IEEEauthorblockA{\textit{Safety Center} \\
\textit{Volvo Car Corporation}\\
Gothenburg, Sweden \\
jian.wu.2@volvocars.com}
\and
\IEEEauthorblockN{Ulrich Sander}
\IEEEauthorblockA{\textit{Safety Center} \\
\textit{Volvo Car Corporation}\\
Gothenburg, Sweden \\
ulrich.sander@volvocars.com}
\and
\IEEEauthorblockN{Carol Flannagan}
\IEEEauthorblockA{\textit{Transportation Research Institute} \\
\textit{University of Michigan}\\
Michigan, USA \\
cacf@umich.edu}
\and
\IEEEauthorblockN{Minxiang Zhao}
\IEEEauthorblockA{\textit{Department of Mechanics and Maritime Sciences} \\
\textit{Chalmers University of Technology}\\
Gothenburg, Sweden \\
minxiang@chalmers.se}
\and
\IEEEauthorblockN{Jonas B\"{a}rgman}
\IEEEauthorblockA{\textit{Department of Mechanics and Maritime Sciences} \\
\textit{Chalmers University of Technology}\\
Gothenburg, Sweden \\
jonas.bargman@chalmers.se}
}

\maketitle

\begin{abstract}
The use of representative pre-crash scenarios is critical for assessing the safety impact of driving automation systems through simulation.
However, a gap remains in the robust evaluation of the similarity between synthetic and real-world pre-crash scenarios and their crash characteristics, such as Delta-v and injury risk.
Without proper validation, it cannot be ensured that the synthetic test scenarios adequately represent real-world driving behaviors and crash characteristics, which may lead to misleading or biased assessments.
One reason for this validation gap is the lack of focus on methods to confirm that the synthetic test scenarios are practically equivalent (or, rather, "similar enough") to real-world ones, given the assessment scope.
Traditional statistical methods, like significance testing, focus on detecting differences rather than establishing equivalence; since failure to detect a difference does not imply equivalence, they are of limited applicability for validating synthetic pre-crash scenarios and crash characteristics.
This study addresses this gap by proposing an equivalence testing method based on the Bayesian Region of Practical Equivalence (ROPE) framework.
This method is designed to assess the practical equivalence of scenario characteristics that are most relevant for the intended assessment, making it particularly appropriate for the domain of virtual safety assessments.
We first review existing equivalence testing methods.
Then we propose and demonstrate the Bayesian ROPE-based method by testing the equivalence of two rear-end pre-crash datasets.
Our approach focuses on the most relevant scenario characteristics, such as key pre-crash kinematics (e.g., the time point at which the following vehicle is not able to avoid a crash) and crash characteristics (e.g., Delta-v and injury risk).
Our analysis provides insights into the practicalities and effectiveness of equivalence testing in synthetic test scenario validation and demonstrates the importance of testing for improving the credibility of synthetic data for automated vehicle safety assessment, as well as the credibility of subsequent safety impact assessments.
\end{abstract}


\section{Introduction}
Driving automation systems \cite{sae_j3016_2021}, including Advanced Driver Assistance Systems (ADAS) \cite{kukkala2018advanced} and Automated Driving Systems (ADS)\cite{chan2017advancements}, show significant promise for reducing crash risks and improving road safety.
Yet, assessing their real-world safety impact remains a challenge \cite{bathla2022autonomous}.

These assessments rely primarily on virtual safety assessments, which offer a cost-effective and efficient alternative to large-scale field tests \cite{brunner2019virtual, riedmaier2020survey}.
In these assessments, baseline test scenarios—pre-crashes without the new technology—are compared to treatment test scenarios that integrate the automation features under assessment \cite{wimmer2023harmonized}.
To yield reliable and statistically sound comparisons, the baseline test scenarios must be comprehensive and accurately represent actual crash conditions \cite{wimmer2023harmonized}.
Unfortunately, the availability of detailed real-world pre-crash data is very limited, typically far below the amount and range required for precise and accurate assessment.
Therefore, synthetic pre-crash scenarios, generated based on real-world data (e.g., using statistical and/or behavioral models), are used as baseline test scenarios \cite{ding2023survey, wu2024modeling, wu2024modelingFV, wu2024generation, Sander2024V4SAFETY}.
The synthetic test scenarios are intended to virtually represent concrete real-world crash scenarios, improve scenario coverage (i.e., to better span the parameter space of the scenario types under assessment), and allow a thorough assessment of a system’s safety impact.

It is important to note that the representativeness of the synthetic pre-crash scenarios, in terms of the characteristics that are relevant to the assessment scope \cite{Sander2024V4SAFETY}, is crucial for an accurate safety impact assessment of a driving automation system.
That is, the synthetic test scenarios must be "similar enough" with respect to the assessment scope.
However, a gap remains in our ability to evaluate this representativeness.
Without proper validation, it cannot be ensured that the synthetic test scenarios adequately represent real-world driving behaviors and crash characteristics, which may lead to misleading or biased assessments \cite{wu2024generation}.

One reason for this validation gap is the lack of focus to date on methods to confirm that the synthetic test scenarios are practically equivalent to real-world ones, given the assessment scope.
Traditional statistical methods, such as significance testing \cite{anderson2000null}, are appropriate for detecting whether differences exist.
However, these methods are less suitable for establishing equivalence, because failure to detect a difference does not imply equivalence \cite{gibbs2013errors}.

This study addresses this gap by proposing an equivalence testing method based on the Bayesian Region of Practical Equivalence (ROPE) \cite{schwaferts2020bayesian}.
The method focuses on characteristics that are most relevant to the intended assessment and assesses whether observed discrepancies are small enough to be practically insignificant instead of testing whether statistically significant differences exist; thus, it is particularly appropriate for validation in the domain of virtual safety assessments.

In the remainder of this paper, Section \ref{section:equivalence testing} outlines existing methods for equivalence testing and emphasizes their potential to improve the formalization of the synthetic test scenario validation process.
Section \ref{section:methodology} details our Bayesian ROPE-based practical equivalence testing method.
Next, Section \ref{section:demonstration} introduces two rear-end pre-crash datasets and applies the proposed ROPE-based method to demonstrate the test for practical equivalence.
Finally, Section \ref{section:discussion} discusses the insights gained regarding the practicality and effectiveness of equivalence testing in validating synthetic test scenarios.

\section{Equivalence Testing} \label{section:equivalence testing}
\begin{table*}[!t]
    \caption{Summary of primary equivalence testing methods}
    \label{tab:equivalence_testing}
    \centering
    \begin{threeparttable}
    \newcolumntype{M}[1]{>{\centering\arraybackslash}m{#1}}
    \begin{tabular}{M{0.75cm}M{1.25cm}M{2.5cm}M{3cm}M{3cm}M{2cm}M{2cm}}
    \hline
    \multicolumn{2}{c}{\multirow{3}{*}{\parbox{2cm}{\vspace*{1.8\baselineskip} \centering Method procedure}}} & \multirow{3}{*}{\parbox{2.5cm}{\vspace*{1.8\baselineskip} \centering Data distribution model}} & \multirow{3}{*}{\parbox{3cm}{\vspace*{2.25\baselineskip} \centering Test scope}} & \multicolumn{3}{c}{\parbox{7cm}{\vspace*{.2\baselineskip} \centering Extend to the multivariate use case}} \\
    \cmidrule(rl){5-7}
    & & & & \multicolumn{2}{c}{Univariate margin/region} & \multirow{2}{*}{\parbox{2cm}{\vspace*{.8\baselineskip} \centering Multivariate region}}\\
    \cmidrule(rl){5-6}
    & & & & A multivariate distance measure & Multiple comparison tests & \\
    \hline
    \multicolumn{2}{c}{\multirow{3}{*}{\parbox{2cm}{\vspace*{2.8\baselineskip} \centering Frequentist TOST-based \cite{schuirmann1981hypothesis}}}} & Parametric & Mean or/and variance & EDNE \cite{hoffelder2015multivariate}, SE \cite{hoffelder2015multivariate} & \multirow{3}{*}{\parbox{2cm}{\vspace*{2.8\baselineskip} \centering \cite{abdi2007bonferroni, lee2018proper}}} & \cite{greene2000claims} \\
    \cmidrule(rl){3-5} \cmidrule(rl){7-7}
    & & \multirow{2}{*}{\parbox{2.5cm}{\vspace*{1.2\baselineskip} \centering Nonparametric$^a$}} & Median (MWW \cite{ahmad1996class, torok2023MWW}) & Generalized MWW \cite{liu2022generalized} & & \multirow{2}{*}{\parbox{2cm}{\vspace*{1.3\baselineskip} \centering $\backslash$}} \\
    \cmidrule(rl){4-5}
    & & & ECDF (KS \cite{massey1951kolmogorov}, AD \cite{scholz1987k}, CVM \cite{medina2024CVMTest})$^b$ & Multivariate KS \cite{fasano1987multidimensional}, AD \cite{zhang2012some}, CVM \cite{cotterill1982limiting} & & \\
    \hline
    \multirow{3}{*}{\parbox{.75cm}{\vspace*{2\baselineskip} \centering Bayesian}} & ROPE-based \cite{schwaferts2020bayesian} & Parametric & Estimated model parameters (e.g., mean, variance, and regression coefficients) & EDNE \cite{hoffelder2015multivariate}, SE \cite{hoffelder2015multivariate} & \cite{abdi2007bonferroni, lee2018proper} & \cite{makowski2019bayestestr} \\
    \cmidrule(rl){2-7}
    & \multirow{2}{*}{\parbox{1.25cm}{\vspace*{0.6\baselineskip} \centering BF-based \cite{kass1995bayes}}} & Parametric & \multirow{2}{*}{\parbox{3.25cm}{ \centering Relative likelihoods of competing models or hypotheses}} & \multicolumn{3}{c}{\multirow{3}{*}{\parbox{7cm}{\centering Directly applicable}}} \\
    \cmidrule(rl){3-3}
    \rule{0pt}{12pt} & & Nonparametric$^b$\\
    \hline
    \end{tabular}
    \begin{tablenotes}
    \RaggedRight
    \item $^a$ These tests were initially designed for testing differences \cite{gibbs2013errors}; thus, they must be adapted for equivalence testing.
    \item $^b$ These methods test the entire distribution.
    \end{tablenotes}
    \end{threeparttable}
\end{table*}

Practical equivalence testing is a set of statistical approaches aimed at determining whether two treatments, processes, or groups are sufficiently similar that any observed differences are small enough to be ignored in practice \cite{limentani2005beyond, greene2000claims}.
This method is crucial in fields such as medicine \cite{greene2000claims, hoffelder2015multivariate}, biology \cite{pallmann2017simultaneous}, and psychology \cite{lakens2018equivalence}, where demonstrating the absence of a meaningful difference is often the primary goal.

Table \ref{tab:equivalence_testing} presents the main methods used in equivalence testing, based on a scoping review we performed.
The methods are categorized based on the statistical procedure, assumptions regarding data distribution, test scope, and dimensionality of the data.
The following describes the table categorization in more detail.

\subsection{Frequentist Approaches}
Traditionally, research has focused on parametric univariate equivalence testing, predominantly employing the frequentist Two One-Sided Tests (TOST) procedure \cite{schuirmann1981hypothesis}.
The TOST method adapts traditional difference tests by performing two one-sided tests to determine whether the difference between specific summary statistics (such as the means or variances) of two groups falls entirely within predefined practical equivalence regions.
If so, the observed differences are considered practically negligible, demonstrating equivalence of the compared groups.
TOST has become the standard procedure for assessing equivalence due to its straightforward application.
It effectively assesses equivalence based on specific summary statistics of two groups, typically assuming normality in the data \cite{limentani2005beyond, greene2000claims, hoffelder2015multivariate, pallmann2017simultaneous}.
While TOST-based parametric methods perform well when their underlying assumptions (such as the data being normally distributed) are met, they can be unreliable when the data deviate from these expectations.

Due to the limitations of parametric methods, nonparametric methods such as the Mann-Whitney-Wilcoxon (MWW) test \cite{ahmad1996class} have gained traction since they do not require the data to follow specific distributions.
It is important to note that the nonparametric methods were originally designed for testing differences \cite{gibbs2013errors}.
These methods typically calculate a (positive) statistic to estimate the difference between two groups and compare it against a predefined margin representing a significant difference.
As a consequence, they must be adapted for equivalence testing; for example, the margin should represent practical equivalence instead.


Moreover, some nonparametric methods can also assess whether entire distributions are equivalent, rather than focusing solely on specific summary statistics like the mean or median.
These methods include the Kolmogorov-Smirnov (KS) test \cite{massey1951kolmogorov}, the Anderson-Darling (AD) test \cite{scholz1987k}, and the Cramér-von Mises (CVM) test \cite{cotterill1982limiting}.
These tests compare empirical cumulative distribution functions (ECDFs) across different groups to evaluate distributional equivalence.

Although these nonparametric methods are robust against violations of parametric assumptions and suitable for testing equivalence in distributions, they may be less powerful than parametric methods when those assumptions hold.

\subsection{Bayesian Approaches}
Within the Bayesian framework, the Region of Practical Equivalence (ROPE) procedure \cite{schwaferts2020bayesian} is commonly used for equivalence testing.
ROPE-based methods assess whether the highest density intervals (HDIs) of the parametric posterior distributions of parameters—such as means, variances, or regression coefficients—fall within predefined practical equivalence regions.

This approach provides a probabilistic interpretation of equivalence while allowing researchers to integrate prior knowledge into their analysis.
By enabling researchers to make more informed decisions regarding the practical equivalence of their findings, it addresses the limitations often encountered in frequentist methods (such as the ambiguous interpretation of p-values and the inability to directly quantify evidence for equivalence) \cite{kruschke2018rejecting}.

Besides ROPE, Bayes factor (BF) \cite{kass1995bayes} can be used for equivalence testing.
Rather than focusing on specific parameters, it compares the relative likelihoods of competing models or hypotheses to determine which is better supported by the data and offers a continuous measure of evidence, facilitating flexible interpretations.
In addition, the BF does not require specific assumptions about the distributions, so it is suitable for both parametric and nonparametric models.
However, a key challenge in using BF for equivalence testing is specifying the competing models—namely, the "equivalence" model and the "difference" model.

Nevertheless, it is important to recognize that both these Bayesian methods, ROPE-based and BF-based, are sensitive to the choice of prior distributions, which can potentially introduce subjectivity into the analysis.

\subsection{Extensions to the Multivariate Use Case}
As research increasingly involves complex, high-dimensional data, there is a growing need to extend equivalence testing methods from univariate to multivariate contexts.
BF-based methods can be directly applied to multivariate data because they compare the likelihood of entire models, naturally accounting for the complexity and interactions among multiple variables \cite{kass1995bayes}.
In contrast, extensions of other methods to the multivariate use case can be categorized into three types: methods that adapt univariate margins or regions to a multivariate framework, those that use multiple comparison tests, and those that define equivalence across entire multivariate regions.

A common strategy for the first category is to create a single, multivariate distance statistic that summarizes differences across multiple variables.
This statistic accounts for correlations between variables and is compared against a predefined equivalence margin.
For example, extensions of TOST-based and ROPE-based parametric methods might employ statistics such as the Euclidean Distance of Nonstandardized Expected values (EDNE) \cite{hoffelder2015multivariate} or Standardized Differences of the Expected (SE) values \cite{hoffelder2015multivariate} to assess equivalence in a multivariate context.
This strategy also applies to TOST-based nonparametric methods.
The Generalized MWW test \cite{liu2022generalized} extends the traditional MWW test to handle multiple outcomes, while multivariate adaptations of ECDF-based methods compare the empirical distributions of multivariate data.
These methods enable a more comprehensive assessment of the equivalence by considering the joint distribution of variables rather than evaluating each variable separately.

The second category of methods is multiple comparison tests (MCTs) \cite{abdi2007bonferroni, lee2018proper}.
In MCTs, the two data groups are declared equivalent only if all tested statistics are within the equivalence regions.
When conducting multiple equivalence tests, the chance of making a false non-equivalence error increases if each test is carried out at the same stringent level as when only one test is performed.
Therefore, adopting less restrictive thresholds for multiple individual tests is advisable.

The third category of methods directly establishes equivalence within a multivariate region.
Instead of assessing equivalence on a variable-by-variable basis, these methods evaluate whether the entire set of multivariate statistics falls within a predefined multivariate region \cite{greene2000claims, makowski2019bayestestr}.
These methods apply only to parametric methods, such as the multivariate ROPE framework \cite{makowski2019bayestestr}, where the equivalence region is defined in a multidimensional space that accounts for correlations among variables.

All three categories facilitate the testing of equivalence in high-dimensional datasets.
However, these method extensions also involve increased computational complexity and potentially lower statistical power, as well as presenting challenges in defining appropriate equivalence regions.
Careful consideration is required to ensure that the chosen method is consistent with the assessment scope.

In summary, Bayesian methods offer a more straightforward interpretation of similarity compared to frequentist methods.
Further, ROPE-based methods naturally integrate domain-specific thresholds without the difficulties in defining competing models that are inherent in BF-based methods.

\section{Methodology} \label{section:methodology}
\begin{figure}[!t]
    \centering
    \includegraphics[width=1\linewidth]{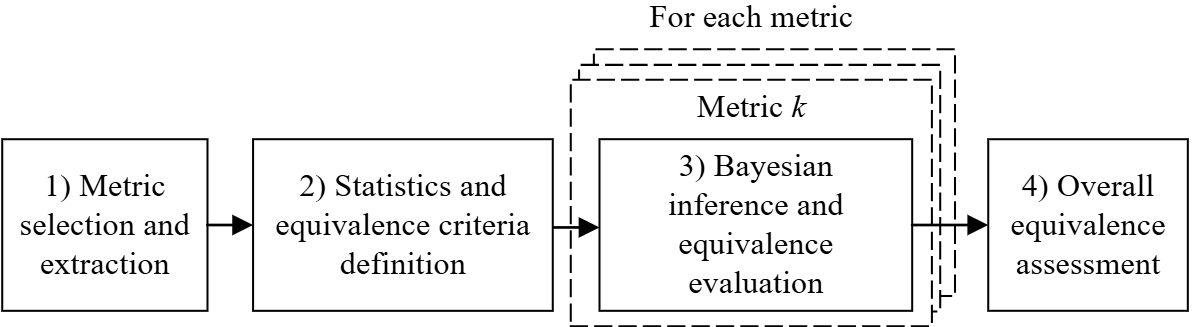}
    \caption{Flowchart of the proposed Bayesian ROPE-based equivalence testing method.}
    \label{fig:methodolgy}
\end{figure}

This study proposes a Bayesian ROPE-based equivalence testing method to robustly evaluate the multivariate similarity between synthetic and real-world (or reference) pre-crash datasets for use in, for example, the validation of baseline test scenario generation.
Fig. \ref{fig:methodolgy} shows the four steps of the method.

\subsection{Step 1: Metric Selection and Extraction}
Pre-crash data typically consist of time series outlining the dynamics and trajectories of the individuals involved in the crash.
Directly comparing these data is often impractical; moreover, the key characteristics relevant to the assessment can differ depending on the specific driving automation systems under investigation.
Hence, the first step is to select the metrics that are most relevant in the assessment and extract/derive them from the pre-crash datasets.

\subsection{Step 2: Statistics and Equivalence Criteria Definition} \label{section:criteria}
For each metric, we choose one or more statistics to assess the differences between the two datasets and establish practical equivalence criteria by specifying both the posterior probability threshold ($\alpha$) and ROPE.
Typically, a 95\% posterior probability threshold is selected, and the definition of ROPE is used to tune the equivalence criteria.
For a statistic to be deemed practically equivalent between two datasets, the 95\% HDI of the statistic's posterior distribution must lie entirely within the defined ROPE.

The chosen statistic(s) and equivalence criteria should accurately reflect the practical equivalence requirements within the assessment scope.
In practice, the choice of statistic(s) should be guided by expert judgment after a thorough analysis of the metric's impact on the performance of the system under assessment.
For instance, metrics of greater relevance or importance may require multiple statistics for a comprehensive assessment.
The KS statistic \cite{lilliefors1967kolmogorov}, used in this work, can be a suitable measure.
It tests the entire distribution by measuring the largest absolute difference between the cumulative distribution functions of the two compared distributions.
(See Sections \ref{section:demonstration} and \ref{section:discussion} for more details.)

\subsection{Step 3: Bayesian Inference and Equivalence Evaluation} \label{section:modeling}
The actions in this step are repeated for each metric.
First, a set of Bayesian distribution models (e.g., exponential, normal, log-normal, gamma, or mixture models) is chosen based on expert judgment and is separately fitted to the reference and synthetic data.
Second, the Widely Applicable Information Criterion (WAIC) \cite{watanabe2010asymptotic} is employed to select the optimal distribution model for each dataset (a lower WAIC indicates better predictive performance).
Third, the chosen statistics are used to measure the difference between the optimal models of both datasets.
Finally, the highest density interval (HDI) of each statistic's posterior distribution is compared with the pre-defined ROPE.
If each HDI is completely within the corresponding ROPE, the data support the practical equivalence of the metric between the compared datasets.

\subsection{Step 4: Overall Equivalence Assessment}
It is crucial to establish clear criteria for overall practical equivalence to summarize all test results with a single overall conclusion.
These criteria should effectively integrate individual test outcomes with expert judgment, ensuring the synthetic dataset adequately represents the real-world conditions that are most relevant to the intended assessment.
The overall equivalence between the datasets is typically recognized if all metrics individually demonstrate equivalence.
However, in practice, a less stringent approach may be justified.
For instance, overall equivalence might be acknowledged if the majority of metrics, explicitly including the most relevant/important ones, demonstrate equivalence, and the remaining metrics do not substantially deviate.
It is important to note that these overall equivalence criteria should be defined beforehand.

Practical equivalence testing relies heavily on expert judgment and interpretive reasoning rather than purely numerical outcomes.
Therefore, the rationale behind declaring equivalence or non-equivalence must be thoroughly documented.
This documentation should clearly outline the motivations behind the selection of specific metrics, statistics, ROPEs, and the overall equivalence criteria.

\section{Demonstration} \label{section:demonstration}
To demonstrate the proposed approach, we performed a (practical) equivalence test between two rear-end pre-crash datasets.
For comparison, we also performed a conventional two-sample KS significance test \cite{lilliefors1967kolmogorov}.

\subsection{Data}
Both datasets contain generated or reconstructed rear-end pre-crash scenarios (i.e., time-series data five seconds before the crash), including the distance between the lead and following vehicles and their longitudinal (along the road) speeds.
These scenarios can, for example, be used as baseline test scenarios for safety impact assessments of driving automation systems.

The first dataset, containing 5,000 pre-crash scenarios, is published online and available to the public \cite{Wu_QUADRIS_project_pre-crash_near-crash}.
These scenarios were generated by modeling real-world rear-end pre-crash data and weighted to accurately represent the population of such crashes with respect to severity in the United States \cite{wu2024modelingFV}.
We consider this first dataset the best representation to date of real-world rear-end crashes across outcome severities (i.e., from only minor physical contact to high severity).

The second dataset consists of 866 reconstructed concrete rear-end pre-crash scenarios from the German In-Depth Accident Study (GIDAS) Pre-Crash Matrix (PCM) dataset \cite{schubert2017gidas}.
This dataset, initiated in 2011, is a subset of the GIDAS dataset that gathers on-scene accident cases with personal injury in Hannover and Dresden, Germany \cite{schubert2017gidas}.
The second dataset is referred to as the "PCM" dataset.

Unlike the first dataset, which encompasses the full severity range with appropriate weighting, the PCM dataset includes only injury-involved crashes, with lower inclusion probabilities for less severe injuries.
To facilitate a fair comparison between the two datasets, we applied an "all-severities-to-GIDAS" transformation \cite{bargman2024methodological} to the first dataset to mimic the GIDAS selection criteria and corresponding censoring.
Consequently, the transform results in a reweighted version of the first dataset that aims to represent injury-involved rear-end crashes in the United States.
The reweighted dataset is here referred to as the "reference" dataset.

\subsection{Metrics and Statistics} \label{section:metrics}
Choosing appropriate metrics and statistics is crucial, since they must capture the scenario characteristics most relevant to the intended assessment.
As noted, these choices should be guided by expert judgment after a thorough analysis of each characteristic's impact on the performance of the system under assessment.

For demonstration purposes, we selected four general metrics that capture crash severity, time available for evasive maneuvers, and vehicle braking behaviors:
\begin{itemize}
    \item $\Delta v_\mathrm{l}\ [\mathrm{m/s}]$: The estimated lead vehicle’s total velocity change (Delta-v) over the crash event.
    \item $t_\mathrm{nr}\ [\mathrm{s}]$: The no-return time, marking the point of no return beyond which a collision is unavoidable even if the following vehicle applies the maximum deceleration of $-9\ \mathrm{m/s^2}$ \cite{zhao2024modeling}.
    Time zero is the impact moment; thus, $t_\mathrm{nr}$ has a negative value.
    \item $a_\mathrm{l,min}\ [\mathrm{m/s^2}]$: The lead vehicle's minimum acceleration (or maximum deceleration).
    \item $a_\mathrm{f,min}\ [\mathrm{m/s^2}]$: The following vehicle's minimum acceleration.
\end{itemize}

$\Delta v_\mathrm{l}$ is a well-established crash severity indicator \cite{shelby2011delta} which is especially important for safety assessment, as high-severity crashes carry a higher risk of serious injuries or fatalities than lower-severity crashes.
We used three statistics to measure the difference between the optimal Bayesian distribution models for the two datasets: 1) the mean difference ($\Delta \overline{\Delta v_\mathrm{l}}$), 2) the KS statistic ($D_{\Delta v_\mathrm{l}}$), and 3) the ratio of high-Delta-v (i.e., $\Delta v_\mathrm{l} \geq$ 15 km/h \cite{linder2003change}) scenario proportions in the two datasets ($\varphi_{\Delta v_\mathrm{l}}$).
Specifically, $\varphi_{\Delta v_\mathrm{l}}$ is computed as
\begin{equation}
    \varphi_{\Delta v_\mathrm{l}} = \eta_{\Delta v_\mathrm{l,pcm}} / \eta_{\Delta v_\mathrm{l,ref}},
\end{equation}
where $\eta_{\Delta v_\mathrm{l,*}}$ is the proportion of high-Delta-v scenarios in dataset $*$.

$t_\mathrm{nr}$ indicates the criticality of the test scenario.
The closer to zero the absolute no-return time ($|t_\mathrm{nr}|$) is, the less critical the situation is.
In test scenarios with a small $|t_\mathrm{nr}|$, the driving automation system under assessment would most likely manage to avoid crashes easily.
Therefore, greater emphasis should be placed on test scenarios with a larger $|t_\mathrm{nr}|$.
As a result,  we used two statistics for $t_\mathrm{nr}$, the KS statistic ($D_{t_\mathrm{nr}}$) and the ratio of high-$|t_\mathrm{nr}|$ scenario proportions ($\varphi_{t_\mathrm{nr}}$, similar to $\varphi_{\Delta v_\mathrm{l}}$), to measure the difference between the optimal Bayesian distribution models for each of the two datasets, regarding only the relevant section (i.e., $|t_\mathrm{nr}| \geq$ 1 s) instead of the entire range.

\begin{figure}[!t]
    \centering
    \subfloat[]{\includegraphics[width=0.22\textwidth]{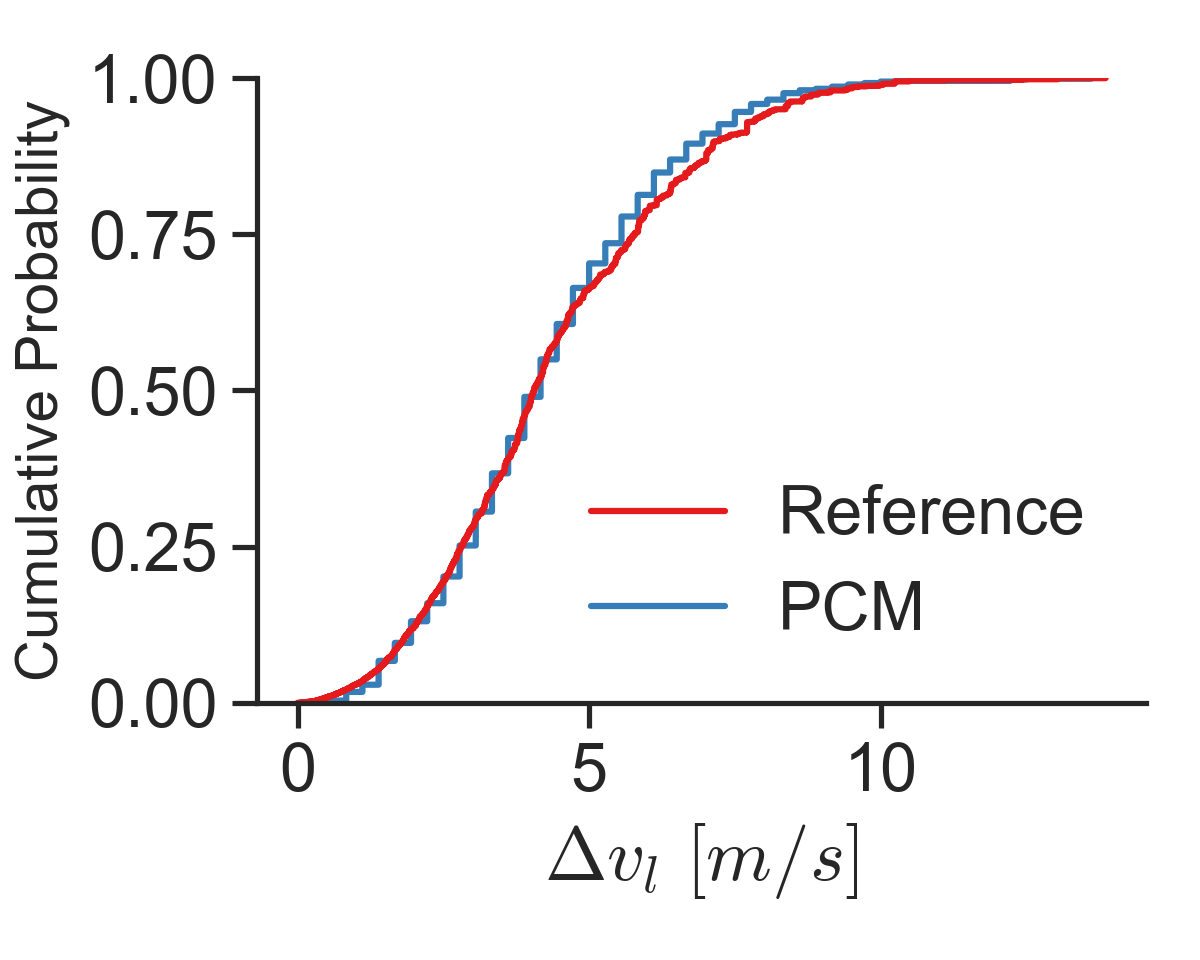}}
    \hfil
    \subfloat[]{\includegraphics[width=0.22\textwidth]{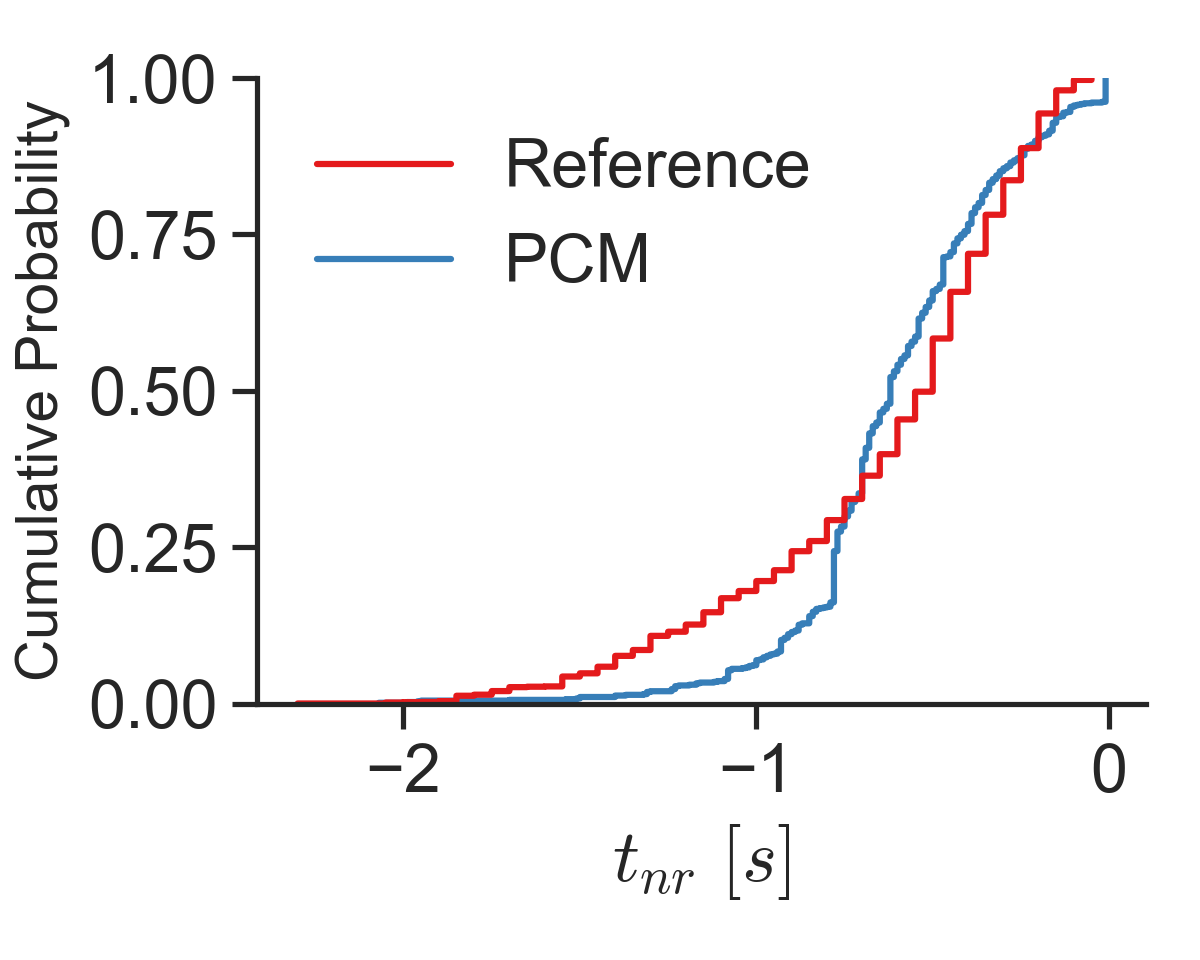}}
    \vfil
    \subfloat[]{\includegraphics[width=0.22\textwidth]{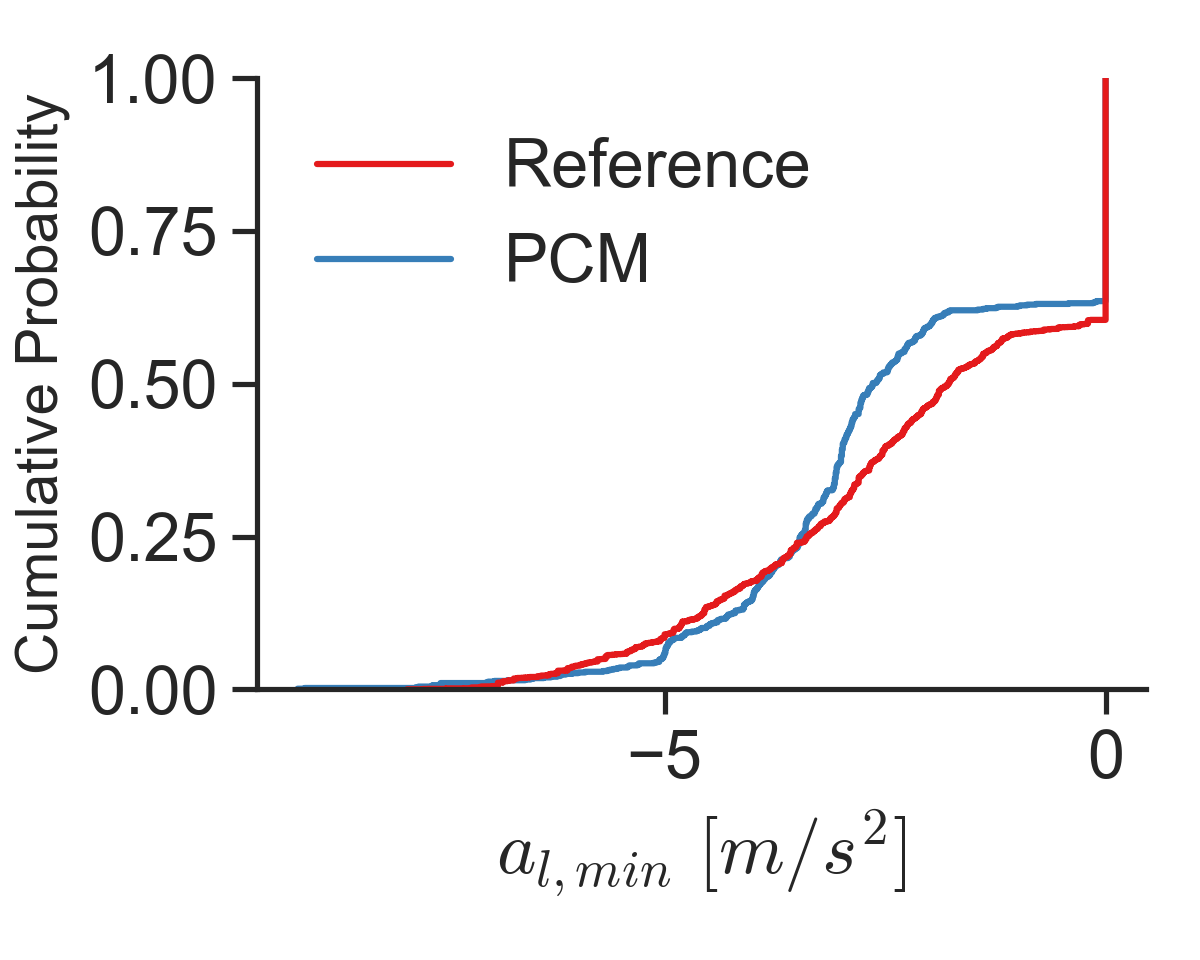}}
    \hfil
    \subfloat[]{\includegraphics[width=0.22\textwidth]{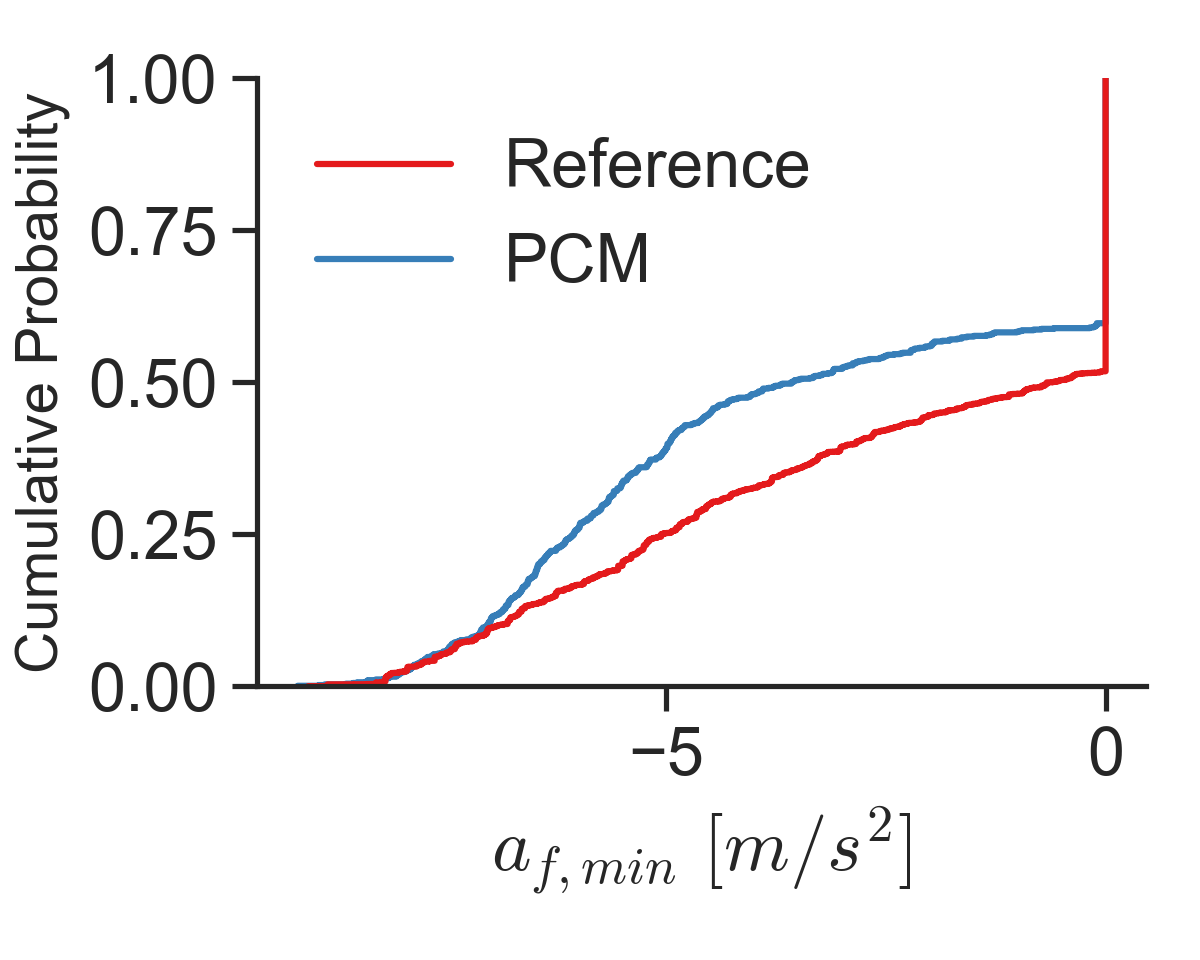}}
    \caption{Weighted cumulative density functions of all four metrics for the reference and PCM datasets: (a) $\Delta v_\mathrm{l}$, (b) $t_\mathrm{nr}$, (c) $a_\mathrm{l,min}$, and (d) $a_\mathrm{f,min}$.}
    \label{fig:cdf}
\end{figure}

The remaining two metrics $a_\mathrm{l,min}$ and $a_\mathrm{f,min}$ describe the braking behaviors of the leading and following vehicles, respectively.
Fig. \ref{fig:cdf} shows the weighted cumulative density functions of all four metrics for the reference and PCM datasets.
Unlike $\Delta v_\mathrm{l}$ and $t_\mathrm{nr}$, $a_\mathrm{l,min}$ and $a_\mathrm{f,min}$ each contain a point mass at 0 m/s$^2$, which indicates two vehicle-behavior patterns (i.e., non-braking and braking).
To measure the patterns separately, we used the following two statistics for each of the two metrics: 1) the ratio of proportions of non-braking (i.e., $a_\mathrm{*,min}$ = 0 m/s$^2$) scenarios ($\varphi_{a_\mathrm{*,min}}$, similar to $\varphi_{\Delta v_\mathrm{l}}$) and 2) the KS statistic for braking (i.e., excluding non-braking cases $a_\mathrm{*,min} < $ 0 m/s$^2$) scenarios ($D_{a_\mathrm{*,min}}$).

\subsection{Equivalence Criteria}
The equivalence criteria, the posterior probability threshold ($\alpha$) and ROPE, determine the practical equivalence and the test's stringency.
They should be specified for each metric test.
In the demonstration, we set $\alpha$ to 0.95 for all statistics and varied the ROPEs (listed in Table \ref{tab:results}) to tune the practical equivalence criteria.
For each statistic, we can conclude its practical equivalence for the two datasets if its 95\% HDI is entirely within the corresponding ROPE.
We further deem a metric practically equivalent only if all its associated statistics demonstrate equivalence.

It is important to note that the ROPEs chosen in this work are purely illustrative and are neither linked to a specific assessment context nor presented as recommendations.
In practice, the ROPEs should be determined through expert judgment after a thorough analysis of each metric’s impact on the performance of the system under assessment; for instance, the ROPE's stringency should increase with the relevance/importance of the corresponding statistic and metric.
On the other hand, as noted, adopting less restrictive ROPEs for a metric with multiple statistics is advisable to mitigate the risk of false non-equivalence conclusions.
In our example, we treated $\Delta v_\mathrm{l}$ as the most critical metric, given its relevance to safety-impact assessments of driving automation systems, followed by $t_\mathrm{nr}$ as the second most important metric.
As a result, we set narrower ROPEs for statistics corresponding to these two metrics (see Table \ref{tab:results}).
For the two less important metrics ($a_\mathrm{l,min}$ and $a_\mathrm{f,min}$), we chose larger ROPEs.

Finally, we also set the overall equivalence criteria: the datasets are deemed practically equivalent if $\Delta v_\mathrm{l}$ and $t_\mathrm{nr}$ meet their equivalence criteria, and at least one of the acceleration metrics meets its equivalence criteria.
Meanwhile, the remaining acceleration metric may deviate slightly, but must have the HDI of each of its statistics contained within a ROPE no more than 1.25 times its original width.

\subsection{Results} \label{section:results}
\begin{table*}[!t]
    \caption{Test results of the reference and PCM datasets}
    \label{tab:results}
    \centering
    \begin{threeparttable}
    \begin{tabular}{cccccccccc}
    \hline
    \multirow{3}{*}{\parbox{1cm}{\vspace*{\baselineskip} \centering Metric}} & \multicolumn{6}{c}{Equivalence test} & \multicolumn{3}{c}{Significance test}\\
    \cmidrule(rl){2-7} \cmidrule(rl){8-10}
    & \multicolumn{2}{c}{Optimal distribution model} & \multirow{2}{*}{Statistic} & \multirow{2}{*}{95\% HDI} & \multirow{2}{*}{ROPE$^a$} & \multirow{2}{*}{Equivalence} & \multirow{2}{*}{KS statistic} & \multirow{2}{*}{P-value} & \multirow{2}{*}{Significance}\\
    \cmidrule(rl){2-3}
    & Reference & PCM \\
    \hline
    \multirow{3}{*}{$\Delta v_\mathrm{l}$} & \multirow{3}{*}{Gamma} & \multirow{3}{*}{Gamma} & $\Delta \overline{\Delta v_l}$ & [-0.11, 0.27] & [-1, 1] & Yes & \multirow{3}{*}{0.06} & \multirow{3}{*}{0.17} & \multirow{3}{*}{No}\\
    & & & $D_{\Delta v_l}$ & [0.01, 0.06] & [0.00, 0.05] & No\\
    & & & $\varphi_{\Delta v_\mathrm{l}}$ & [0.91, 1.06] & [0.95, 1.05] & No\\
    \cmidrule(rl){1-10}
    \multirow{2}{*}{$t_\mathrm{nr}$} & \multirow{2}{*}{Gamma} & \multirow{2}{*}{Gamma} & $D_{t_\mathrm{nr}}$ & [0.00, 0.06] & [0.00, 0.10] & Yes & \multirow{2}{*}{0.15} & \multirow{2}{*}{0.00} & \multirow{2}{*}{Yes$^d$}\\
    & & & $\varphi_{t_\mathrm{nr}}$ & [0.68, 0.95] & [0.90, 1.10] & No\\
    \cmidrule(rl){1-10}
    \multirow{2}{*}{$a_\mathrm{l,min}$} & \multirow{2}{*}{Mixture$^b$} & \multirow{2}{*}{Mixture$^c$} & $D_{a_\mathrm{l,min}}$ & [0.09, 0.14] & [0.00, 0.15] & Yes & \multirow{2}{*}{0.14} & \multirow{2}{*}{0.00} & \multirow{2}{*}{Yes$^d$}\\
    & & & $\varphi_{a_\mathrm{l,min}}$ & [0.81, 1.02] & [0.80, 1.20] & Yes\\
    \cmidrule(rl){1-10}
    \multirow{2}{*}{$a_\mathrm{f,min}$} & \multirow{2}{*}{Mixture$^b$} & \multirow{2}{*}{Mixture$^b$} & $D_{a_\mathrm{f,min}}$ & [0.13, 0.22] & [0.00, 0.15] & No & \multirow{2}{*}{0.16} & \multirow{2}{*}{0.00} & \multirow{2}{*}{Yes$^d$}\\
    & & & $\varphi_{a_\mathrm{f,min}}$ & [0.75, 0.92] & [0.80, 1.20] & No\\
    \hline
    \end{tabular}
    \begin{tablenotes}
    \RaggedRight
    \item $^a$ \textbf{The ROPEs are purely illustrative and are neither linked to a specific assessment context nor presented as recommendations.}
    \item $^b$ A mixture distribution model that combines a binomial distribution and a truncated normal distribution.
    \item $^c$ A mixture distribution model that combines a binomial distribution and a gamma distribution.
    \item $^d$ The difference is significant at the 0.05 significance level.
    \end{tablenotes}
    \end{threeparttable}
\end{table*}

As noted, $a_\mathrm{l,min}$ and $a_\mathrm{f,min}$ contain a point mass at 0 m/s$^2$, which requires a mixture distribution model to describe its distribution \cite{wu2024modeling}.
Here, the mixture distribution model combines two distributions: all point mass values come from a binomial distribution, while all other values are assumed to be from a continuous distribution.
As described in Section \ref{section:modeling}, a set of Bayesian distribution models was fitted to both datasets.
Then, the models with the lowest WAIC were selected.
Specifically, for the metrics without a point mass (i.e., $\Delta v_\mathrm{l}$ and $t_\mathrm{nr}$), exponential, gamma, normal, and log-normal distribution models were fitted.
For the metrics containing a point mass (i.e., $a_\mathrm{l,min}$ and $a_\mathrm{f,min}$), the fitted models were mixture distribution models: the continuous distribution could be an exponential, gamma, log-normal, or truncated normal distribution.

Table \ref{tab:results} shows the test results of the practical equivalence and statistical significance tests for the reference and PCM datasets.
For each metric, the optimal Bayesian distribution models for both datasets are listed.

For the equivalence tests, all the $a_\mathrm{l,min}$ statistics passed the equivalence tests, indicating the practical equivalence of this metric.
However, the results show non-equivalence of the other three metrics.
Meanwhile, all significant tests except the one for $\Delta v_\mathrm{l}$ demonstrated a statistically significant difference at the 0.05 significance level.
The comparison of the equivalence and significance test results reveals no obvious connection: a statistically nonsignificant difference does not imply practical equivalence (as seen in the test results for $\Delta v_\mathrm{l}$), while a statistically significant difference may still be practically negligible (as seen in the test results for $a_\mathrm{l,min}$).
In summary, given the results that only one metric ($a_\mathrm{f,min}$), a less-critical one, passed the equivalence test while the other metrics failed, the demonstration indicates that the two datasets are not considered practically equivalent.

However, as noted, the ROPEs chosen here are purely illustrative.
For instance, we set stringent ROPEs for Delta-v related statistics just to show the conflict results between equivalence and significance tests.
In practice, more relaxed ROPEs should probably be chosen instead. 
The purpose of the demonstration is to showcase the procedure of the proposed Bayesian ROPE-based equivalence testing method, making the specific results less relevant.


\section{Discussion and conclusions} \label{section:discussion}
This study proposes a Bayesian method for equivalence testing based on ROPE to validate synthetic pre-crash scenarios against real-world ones.
By defining clear ROPE-based practical equivalence criteria, our method advances beyond traditional significance tests, enabling an evaluation of whether two datasets are "similar enough" for practical safety assessments.
Unlike many standard ROPE-based methods that rely on the assumption of normally distributed data, the proposed approach fits a family of Bayesian distribution models and chooses the best-fitting one.
With this strategy, the method effectively handles mixture distributions, such as those with point masses.
Our method is also flexible enough to be able to use different statistics tailored to individual metrics, enabling more precise tuning of practical equivalence criteria for specific applications than is typically supported by conventional methods.
For instance, the method permits combining statistics commonly used in nonparametric tests (e.g., the KS statistic) with summary statistics (e.g., differences in means, medians, or proportions), according to their relevance and importance in the context of the safety assessment.

Of note, several key practical considerations exist when conducting an equivalence test.
First, it is crucial to select the right metrics: they must represent the scenario characteristics that are most relevant for the intended assessment.
Using more metrics adds comparison dimensions, making the test more thorough while also increasing the chance of making a false non-equivalence error.
Therefore, choosing only the most relevant metrics for the assessment scope is essential, as it helps avoid mistakenly declaring non-equivalence due to non-equivalent, less-relevant/important metrics.

Second, choosing appropriate statistics and defining the corresponding ROPEs for each metric must be done carefully, as these choices dictate both the scenario characteristics being measured and the stringency of those measurements.
The goal is to ensure that the chosen statistics and ROPEs collectively capture the practical equivalence requirements relevant to the assessment.
As mentioned, these selections should be guided by expert judgment, based on a thorough analysis of each metric's impact on the performance of the system under assessment and the number of associated statistics.
In our demonstration, we used the KS statistic, which measures the largest absolute difference between the cumulative distribution functions of two datasets, applying uniform importance across the entire metric range.
However, real-world applications may benefit from a sensitivity analysis to identify which intervals of the metric have the greatest impact on the assessment of a particular system, enabling a metric- and statistic-dependent ROPE that is more stringent for more sensitive intervals.

Finally, deciding the overall equivalence involves integrating the test outcomes from all selected statistics and metrics into a coherent conclusion.
A structured decision rule should be established beforehand, clearly outlining how the results from individual tests will contribute to the final judgment of equivalence.
Specifying criteria for situations where partial equivalence occurs is beneficial, ensuring transparency and robustness in the final equivalence determination.

In practice, if the datasets fail to meet the overall equivalence criteria, reviewing and adjusting the synthetic test scenario generation process may be necessary.
Alternatively, one could reweight the synthetic dataset using methods like post-stratification \cite{holt1979post} and iterative proportional fitting \cite{choupani2016population}, aligning the distributions of metrics that did not meet the equivalence criteria.

A key limitation of this study is that the metrics were modeled and evaluated independently, ignoring potential correlations between them.
Real-world crash scenario metrics, such as crash severity ($\Delta v_\mathrm{l}$) and temporal criticality ($t_\mathrm{nr}$), often exhibit inherent correlations due to their shared underlying dynamics.
Treating these metrics independently could result in overly conservative practical equivalence decisions and a higher incidence of false non-equivalence conclusions.
Future work should address this limitation by employing multivariate Bayesian ROPE-based equivalence testing methods.
Such methods explicitly consider the joint distributions and interactions among metrics, enhancing the accuracy and reliability of equivalence assessments.

In addition, the sensitivity of the ROPE method to prior assumptions is important, especially in regulatory contexts where consistency and transparency are essential.
Future studies should conduct explicit sensitivity analyses to quantify how the choice of priors affects equivalence conclusions.

Despite these limitations, the proposed method is an important initial step toward a systematic and transparent practical equivalence test between synthetic and real-world pre-crash datasets.
In fact, the method provides practitioners with a systematic and rigorous tool to validate synthetic scenario generation processes by clearly quantifying similarities and discrepancies between synthetic and real-world datasets.
It enables targeted identification of aspects needing refinement, thus guiding iterative improvements in scenario realism and representativeness.

\section*{Acknowledgment}
This research was funded by FFI Vinnova, a Swedish governmental agency for innovation, as part of the project Improved quantitative driver behavior models and safety assessment methods for ADAS and AD (QUADRIS: nr. 2020-05156).


\bibliographystyle{IEEEtran}
\bibliography{references}

\end{document}